%% file: main.tex
\pdfoutput=1
\documentclass[11pt]{article}

\usepackage[letterpaper, margin=1in]{geometry}
\usepackage[T1]{fontenc}
\usepackage[utf8]{inputenc}
\usepackage{lmodern}
\usepackage{microtype}

\usepackage{amsmath,amssymb,amsfonts}
\usepackage{mathtools}
\usepackage{bm}
\usepackage{amsthm}

\usepackage{graphicx}
\usepackage{xcolor}
\usepackage{booktabs}
\usepackage[font=small, labelfont=bf]{caption}
\usepackage{subcaption}
\usepackage{tikz}
\usetikzlibrary{arrows, arrows.meta, calc, positioning}
\def\plotlinewidth{1pt}

\usepackage{cite}
\usepackage{siunitx}
\usepackage{optidef}
\usepackage{url}
\usepackage{varwidth}
\usepackage{stackengine}
\usepackage[colorlinks=true, allcolors={blue!55!black}]{hyperref}

\newcommand\barbelow[1]{\stackunder[1.2pt]{$#1$}{\rule{.8ex}{.075ex}}}
\newcommand\Umbruch[2][3cm]{\begin{varwidth}{#1}\centering#2\end{varwidth}}

\makeatletter
\def\bstctlcite{\@ifnextchar[{\@bstctlcite}{\@bstctlcite[@auxout]}}
\def\@bstctlcite[#1]#2{\@bsphack
  \@for\@citeb:=#2\do{%
    \edef\@citeb{\expandafter\@iden\@citeb}%
    \if@filesw\immediate\write\csname #1\endcsname{\string\citation{\@citeb}}\fi}%
  \@esphack}
\makeatother

\input{00_template/commands.tex}
\input{00_template/glossary.tex}
\glsdisablehyper
\input{00_template/symbols.tex}

\newenvironment{keywords}%
  {\begin{quote}\small\noindent\textbf{Keywords:}\ }%
  {\end{quote}}

\title{Time-Optimal Cooperative Manipulation of Heavy Objects\\
  with Dual-Arm Robots}

\author{%
  Maximilian Dio \qquad Knut Graichen \qquad Andreas V\"olz\\[1.5ex]
  \small Chair of Automatic Control, Friedrich-Alexander-Universit\"at Erlangen-N\"urnberg\\
  \small Cauerstra\ss e 7, 91058 Erlangen, Germany\\
  \small\texttt{\{maximilian.dio, knut.graichen, andreas.voelz\}@fau.de}%
}

\date{}

\begin{document}

\bstctlcite{IEEEexample:BSTcontrol}

\maketitle

\input{01_content/abstract.tex}


\input{01_content/introduction.tex}

\input{01_content/related_work.tex}

\input{01_content/model_description.tex}

\input{01_content/problem/problem_formulation.tex}

\input{01_content/experimental_evaluation.tex}

\input{01_content/conclusion.tex}

\section*{Acknowledgments}
This work was funded by the Deutsche Forschungsgemeinschaft (DFG, German Research Foundation)~-- 438833210; 493539176.

\bibliographystyle{IEEEtran}
\bibliography{IEEEabrv,../00_template/library}

\end{document}

%% file: 00_template/commands.tex
\newcommand{\secref}[1]{Section~\ref{#1}}
\newcommand{\figref}[1]{Fig.~\ref{#1}}
\newcommand{\tabref}[1]{Tab.~\ref{#1}}

%% file: 00_template/glossary.tex
\usepackage[acronym]{glossaries}

\newacronym{hqp}{HQP}{hierarchical quadratic programming}
\newacronym{qp}{QP}{quadratic program}
\newacronym{lp}{LP}{linear program}
\newacronym[plural=DOF,firstplural=degrees of freedom (DOFs)]{dof}{DOF}{degree of freedom}
\newacronym{lqr}{LQR}{linear quadratic regulator}
\newacronym{com}{COM}{center of mass}
\newacronym{pidc}{PIDC}{projected inverse dynamics control}
\newacronym{ode}{ODE}{ordinary differential equation}
\newacronym{dae}{DAE}{differential algebraic equation}
\newacronym{pd}{PD}{proportional derivative}
\newacronym{pid}{PID}{proportional derivative integral}
\newacronym{fri}{FRI}{Fast Robot Interface}
\newacronym{ros2}{ROS2}{Robot Operating System 2}
\newacronym{co}{CO}{convex optimization}
\newacronym{ni}{NI}{numerical integration}
\newacronym{dp}{DP}{dynamic programming}
\newacronym{ra}{RA}{reachability analysis}
\newacronym{topp}{TOPP}{time-optimal path parameterization}
\newacronym{topp-ra}{TOPP-RA}{time-optimal path parameterization with reachability analysis}
\newacronym{topp3-ra}{TOPP3-RA}{third-order time-optimal path parameterization via reachability analysis}
\newacronym{topp-nlp}{TOPP-NLP}{time-optimal path parameterization with nonlinear programming}
\newacronym{topp-ra-nlp}{TOPP-RA-NLP}{time-optimal path parameterization with reachability analysis and nonlinear programming}
\newacronym{rhtopp}{RHTOTOPP}{receding horizon time-optimal path parameterization}
\newacronym{rhtopp-ra}{RH-TOPP-RA}{receding horizon time-optimal path parameterization via reachability analysis}
\newacronym[shortplural={MARS}]{mars}{MARS}{multi-arm robotic system}
\newacronym{ocp}{OCP}{optimal control problem}
\newacronym{pmp}{PMP}{Pontryagin's maximum principle}
\newacronym{socp}{SOCP}{second-order cone program}
\newacronym{ik}{IK}{inverse kinematics}
\newacronym{mpc}{MPC}{model predictive control}
\newacronym{mvc}{MVC}{maximum velocity curve}
\newacronym{mocap}{MoCap}{motion capture}
\newacronym{vmrs}{VMRS}{Volume Maximizing Ray Shooting}
\newacronym{nlp}{NLP}{non-linear program}
\newacronym{rh}{RH}{receding horizon}
\newacronym{rha}{RHA}{receding horizon approach}
\newacronym{slsqp}{SLSQP}{sequential least squares programming}
\newacronym{emf}{EMF}{electromotive force}
\newacronym{otg}{OTG}{online trajectory generation}
\newacronym{mppi}{MPPI}{model predictive path integral}
\newacronym[plural=RMPs,firstplural=Riemannian Motion Policies (RMPs)]{rmp}{RMP}{Riemannian Motion Policy}
\newacronym{ics}{ICS}{inevitable collision state}
\newacronym{ds}{DS}{dynamical system}
\newacronym[plural = CBFs, firstplural = control barrier functions (CBFs)]{cbf}{CBF}{control barrier function}

%% file: 00_template/symbols.tex
\usepackage{amsmath}
\usepackage{amssymb}

\newcommand{\quatSymbol}{\varphi}

\newcommand{\posSymbol}{p}
\newcommand{\vLSymbol}{v}
\newcommand{\vASymbol}{\omega}
\newcommand{\twistSymbol}{\nu}
\newcommand{\Null}{\boldsymbol{\mathcal{N}}}

\newcommand{\eye}{\boldsymbol{I}}
\newcommand{\zM}{\boldsymbol{0}}
\newcommand{\zV}{\boldsymbol{0}}

\newcommand*{\vm}[1]{\boldsymbol{#1}}

\newcommand*{\pinv}{^{\#}}
\newcommand*{\skewvec}[1]{\left[ #1 \right]_{\times}}

\newcommand{\T}{\top}

\newcommand{\idx}{i\in\{1,2\}}

\newcommand*{\st}{\operatorname{s.\!t.}}

\newcommand{\SE}{\textit{SE}}

\newcommand{\SEn}[1]{\SE(#1)}

\newcommand{\Rn}[1]{\mathbb{R}^{#1}}
\newcommand{\Hn}[1]{\mathbb{H}^{#1}}

\newcommand*{\reference}{^{\mathrm{ref}}}
\newcommand*{\virtual}{^{\mathrm{virt}}}
\newcommand*{\internal}{_{\mathrm{int}}}
\newcommand*{\cycle}{_{\mathrm{cyc}}} %

\newcommand{\vnorm}[3]{\left\lVert {#1} \right\lVert^{#2}_{#3}}

\newcommand{\diff}{\mathrm{d}}

\newcommand{\Rotm}{\boldsymbol{R}}
\newcommand{\Transform}{\boldsymbol{T}}
\newcommand{\pos}{\boldsymbol{\posSymbol}}
\newcommand{\quat}{\boldsymbol{\quatSymbol}}

\newcommand{\rvec}{\boldsymbol{r}}

\newcommand{\twist}{\boldsymbol{\twistSymbol}}
\newcommand{\vL}{\boldsymbol{\vLSymbol}}
\newcommand{\vA}{\boldsymbol{\vASymbol}}

\newcommand{\wrench}{\boldsymbol{h}}

\newcommand{\M}{\boldsymbol{M}}

\newcommand{\n}{\boldsymbol{c}}

\newcommand{\g}{\boldsymbol{g}}
\newcommand{\G}{\boldsymbol{G}}
\newcommand{\N}{\Null}

\newcommand{\constraintmatrix}{\boldsymbol{\Lambda}}
\newcommand{\constraintvector}{\boldsymbol{\beta}}

\newcommand{\inputmatrix}{\boldsymbol{B}}
\newcommand{\J}{\vm{J}}

\newcommand{\jt}{\boldsymbol{\tau}}

\newcommand{\s}{s}
\newcommand{\cSet}{\mathcal{K}}
\newcommand{\as}{\Omega}

\newcommand{\fSet}{\Omega}

\newcommand{\xPath}{\boldsymbol{x}}
\newcommand{\uPath}{\boldsymbol{u}}

\newcommand{\dtwist}{\vm{\dot{\twistSymbol}}}
\newcommand{\ddtwist}{\vm{\ddot{\twistSymbol}}}

\newcommand{\hatJ}{\vm{\hat{J}}}

\newcommand{\q}{\boldsymbol{q}}
\newcommand{\dq}[1]{\vm{\dot{q}}_{#1}}
\newcommand{\ddq}[1]{\vm{\ddot{q}}_{#1}}

\newcommand{\hatjt}{\vm{\hat{\tau}}}

\newcommand{\ds}{\dot{s}}
\newcommand{\dds}{\ddot{s}}
\newcommand{\ddds}{s^{(3)}}

%% file: 01_content/abstract.tex
\begin{abstract}
Cooperative dual-arm manipulation enables robots to handle heavy objects beyond individual arm capacities, but time-optimal motion under kinematic, dynamic, and jerk constraints is challenging due to actuation redundancy and the risk of excessive internal wrenches damaging the workpiece.
This paper presents an integrated trajectory planning and control framework that addresses these challenges.
The \gls{topp3-ra} method is extended to redundantly actuated systems by incorporating internal wrenches as optimization variables.
It reduces the problem dimensionality compared to existing formulations, improves efficiency, and allows linear complexity in the number of discretization points.
A \gls{hqp} controller resolves the remaining wrench redundancy online, enforcing joint torque limits and bounding internal wrenches during trajectory tracking.
The framework is validated experimentally on two KUKA LBR iiwa 7 R800 manipulators cooperatively manipulating a \SI{16}{\kilogram} payload, exceeding each arm's \SI{7}{\kilogram} nominal capacity by more than twofold.
The system achieves position tracking errors below \SI{6}{\milli\meter} and orientation errors below \SI{1.01}{\degree} while satisfying all velocity, acceleration, jerk, and torque constraints in real time.
\end{abstract}

%% file: 01_content/introduction.tex
\section{Introduction}

Collaborative manipulators are increasingly deployed in human-centered environments due to their compact size and safety features \cite{Sherwani2020}.
However, their limited manipulation capacity of heavy or bulky objects remains a major drawback \cite{Matheson2019}, motivating the use of  cooperative multi-arm and humanoid robotic systems \cite{SinaMirrazaviSalehian2016}.
In industrial applications, cycle time directly impacts throughput.
Minimizing execution time while respecting physical constraints is therefore essential for deploying cooperative systems in practice.

Cooperative manipulation presents substantial challenges. Kinematic loops between the arms and the grasped object give rise to internal wrenches that may damage workpieces.
Redundant actuation further complicates the problem, as wrench distribution among the arms is not unique.
Ensuring feasible control commands that prevent torque violations and excessive internal wrenches is critical \cite{Dio2023}.

Efficient operation requires time-optimal motion planning under kinematic and dynamic constraints.
Decoupled trajectory planning, i.e., first optimizing the path and subsequent timing computation simplifies the problem. The latter step is formulated as the \gls{topp} problem.
To prevent oscillations and damage, smooth trajectories are essential, which can be achieved by constraining jerk \cite{Reiter2018}.
Due to the inherent actuation redundancy in coupled manipulation, post-processing is necessary to uniquely determine torque and wrench trajectories \cite{Dio2024}.

\subsection{Contributions}

This paper presents an integrated framework for time-optimal cooperative dual-arm manipulation by extending trajectory optimization and control methods to redundantly actuated systems.
The main contributions are:
\begin{itemize}
    \item Extension of the \gls{topp3-ra} method \cite{Dio2025} to redundantly actuated dual-arm systems through a problem reformulation that treats internal wrenches as optimization variables.
    \item Adaptation of the \gls{hqp} control strategy \cite{Dio2023} to incorporate the resulting time-optimal reference trajectories, forming a combined planning and control framework in which the unresolved wrench redundancy from the planner is handled online by the controller.
    \item Experimental validation on a real dual-arm system, demonstrating manipulation of payloads exceeding nominal load capacities.
\end{itemize}

\subsection{Paper Organization}

The remainder of this paper is organized as follows.
\secref{sec:related_work} reviews related work on time-optimal trajectory generation and cooperative dual-arm control.
\secref{sec:model_description} presents the coupled dual-arm system model.
The \gls{topp} formulation is described in \secref{sec:topp}, followed by the \gls{hqp} control architecture in \secref{sec:hqp}.
\secref{sec:experimental_evaluation} presents experimental results.
\secref{sec:conclusion} concludes the paper.

%% file: 01_content/related_work.tex
\section{Related Work}\label{sec:related_work}

Assuming a predefined geometric path, \gls{topp} aims to find the time-optimal velocity profile under kinematic and dynamic constraints \cite{Bobrow1985}.
\Gls{ni} methods compute trajectories by integrating maximum and minimum accelerations from the boundary states, but require identifying switching points \cite{Bobrow1985, Pham2014}. Extensions handle redundant actuation \cite{Liu1991, PhamStasse2015}, and third-order constraints \cite{PhamTOPP32017}.

Convex formulations in the path domain enable efficient solutions \cite{Verscheure2008}. An adaptation to third-order constraints is achieved via sequential convex programming \cite{Debrouwere2013}. In \cite{Ji2024}, a convex restriction approach handles jerk constraints via B-spline parameterization of the velocity profile. In \cite{Dio2024} sequential linear programming is used to obtain time-optimal trajectories for redundantly actuated manipulators, ensuring smoothness by constraining jerk and torque rate.

The authors of \cite{Huang2023} propose a sampling-based method that constructs a tree of feasible nodes to find time-optimal trajectories with jerk constraints.
A benchmark study in \cite{Lee2024} systematically evaluates the performance of jerk-constrained methods on industrial manipulators.
Dynamic programming approaches exist for third-order constraints but are computationally expensive \cite{Kaserer2019, Kaserer2020}.

\Gls{ra} methods offer simpler implementations and superior computational complexity compared to convex optimization \cite{PhamRA2017}.
In \cite{Dio2025}, the method was augmented to handle third-order constraints for non-redundant manipulators. However, extending \gls{topp3-ra} to redundantly actuated systems is still an open problem, which this paper addresses by reformulating the problem to treat internal wrenches as optimization variables.

Cooperative manipulation extends single-arm interaction control by simultaneously controlling motion and interaction forces between robots \cite{Smith2012, Abbas2023}.
The operational space formulation \cite{Khatib1987} provides the theoretical basis for task-space motion and force control, while the virtual linkage model \cite{Williams1993} offers a physical characterization of internal forces in multi-grasp scenarios.
These systems are redundantly actuated, requiring wrench distribution through grasp matrix pseudo-inverses to prevent internal stresses \cite{Walker1991, Chung2005, Erhart2015}.

Classical control approaches include primary-replica control \cite{KIKUCHI19842349}, hybrid position/force control \cite{Uchiyama1988}, and impedance-based methods building on \cite{Hogan1985, Caccavale2008, Erhart2016}.
The authors of \cite{Verginis2023} present a rigidity theory-based approach for cooperative manipulation, focusing on internal force regulation to ensure stable grasping and coordinated motion.
Constraint enforcement via online optimization has emerged as essential when working at payload limits as demonstrated in \cite{Dio2023} which employs a \gls{hqp} framework for dual-arm torque control under strict constraints.
Task-priority frameworks \cite{Nakamura1987, Siciliano1991} and their extension to whole-body control \cite{Sentis2005} laid the groundwork for modern \gls{hqp} formulations, which generalize the prioritized control to inequality constraints \cite{Kanoun2011} and effectively handle multiple priorities and hard constraints like torque limits \cite{Escande2014}.

While existing methods address time-optimality and cooperative control separately, an integrated framework for time-optimal planning and control of cooperative dual-arm manipulation of heavy objects that resolves actuation redundancy while enforcing physical constraints is still missing.
This paper addresses this gap by integrating \gls{topp} for redundantly actuated systems with a \gls{hqp} control strategy.

%% file: 01_content/model_description.tex
\section{System Modeling} \label{sec:model_description}

\begin{figure}[tb]
    \centering
    \includegraphics[width=0.85\textwidth]{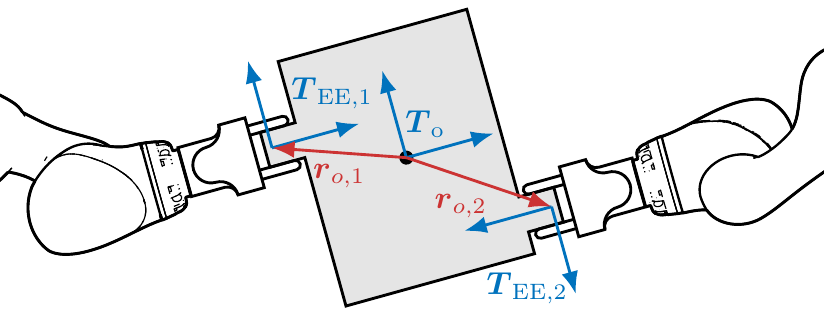}
    \caption{Cooperative manipulation with dual-arm robot. Coordinate frames of end effectors and object.}
     \label{fig:coupled_kinematics}
\end{figure}

Consider an object being manipulated by two robotic manipulators as illustrated in Fig.~\ref{fig:coupled_kinematics}.
The object pose is represented by $\Transform_o \in \SEn{3}$ which is equivalently represented by $\q_o^\T =\begin{bmatrix} \pos_o^\T & \quat_o^\T \end{bmatrix} \equiv \Transform_o$.
Here, $\pos_o \in \Rn{3}$ is a position and a unit quaternion $\quat_o \in \Hn{}$ describes the orientation.
Its twist $\twist_o^\T = \begin{bmatrix} \vL_o^\T & \vA_o^\T \end{bmatrix} $ is given as the concatenation of linear velocities $\vL_o \in \Rn{3}$ and angular velocities $\vA_o \in \Rn{3}$.
Furthermore, consider two fixed-base $n$-\gls{dof} manipulators with joint configurations $\q_i \in \Rn{n}$ for $i \in \{1,2\}$.
The augmented system configuration is $\q^\T = \begin{bmatrix} \q_o^\T & \q_1^\T & \q_2^\T \end{bmatrix}$.
The corresponding system velocity is $\twist^\T = \begin{bmatrix} \twist_o^\T & \dq{1}^\T & \dq{2}^\T \end{bmatrix}$, comprising the object's twist and the joint velocities.

The manipulators are coupled via a rigid grasp on the object as shown in Fig.~\ref{fig:coupled_kinematics}.
This requires the end effector poses $\Transform_{\mathrm{EE,i}}(\q_i)$ to be equal to grasp poses $\Transform_{\mathrm{o,i}}(\q_o)$ at the object, i.e.,
\begin{equation}\label{eq:pose_constraints}
    \Transform_{\mathrm{EE,i}}(\q_i) = \Transform_{\mathrm{o,i}}(\q_o), \quad \idx \,.
\end{equation}

To enforce coupling between the object and the end effectors, a constrained optimization problem can be defined as mentioned in  \cite{carpentier:hal-03271811}. A similar procedure is followed here. The resulting constrained dynamics of the object is given by
\begin{equation} \label{eq:obj_dyn_augmented}
    \M_o(\q_o) \dtwist_o + \n_o(\q_o,\twist_o) + \g_o(\q_o) - \G(\q_o) \wrench = \zV \,,
\end{equation}
with inertia matrix $\M_o(\q_o)$, Coriolis effects $\n_o(\q_o,\twist_o)$, and gravitational forces $\g_o(\q_o)$.
The wrenches $\wrench^\T = \begin{bmatrix} \wrench_1^\T & \wrench_2^\T \end{bmatrix}$ are projected onto the object \gls{com} via the grasp matrix $\G := \G(\q_o)= \begin{bmatrix} \G_1(\q_o) & \G_2(\q_o) \end{bmatrix}$, where each partial grasp matrix is defined as
\begin{equation}
    \G_i(\q_o) = \begin{bmatrix}\eye_3 & \zM_3 \\ \skewvec{\rvec_{\mathrm{o,i}}(\q_o)}  & \eye_3 \end{bmatrix} \,.
\end{equation}
The vector $\rvec_{\mathrm{o,i}}$ defines the relative position between the object's \gls{com} and the grasp frame and $\skewvec{\cdot}$ denotes the skew-symmetric matrix operator.

Similarly, the constrained manipulator dynamics are
\begin{equation} \label{eq:man_dyn_augmented}
    \M_i(\q_i) \ddq{i} + \n_i(\q_i,\dq{i}) + \g_i(\q_i) + \J_i(\q_i)^\T \wrench_i = \jt_i\,,
\end{equation}
for $\idx$ with generalized mass matrix $\M_i(\q_i)$. The generalized Coriolis and centrifugal terms are $\n_i(\q_i,\dq{i})$ and the generalized gravitational forces are $\g_i(\q_i)$. The interaction wrenches $\wrench_i$ act on the end effectors and are projected onto the joint torques $\jt_i$ via transposed end-effector Jacobian $\J_i(\q_i)^\T$.

The resulting constrained system dynamics are
\begin{align} \label{eq:system}
    \begin{split}
    & \underbrace{\begin{bmatrix}
            \M_o & \zM & \zM \\ \zM & \M_1 & \zM \\ \zM & \zM & \M_2
        \end{bmatrix}}_{\M(\q)} \dtwist + \underbrace{\begin{bmatrix}
            \n_o \\ \n_1 \\ \n_2
        \end{bmatrix}}_{\n(\q,\twist)}
        + \underbrace{\begin{bmatrix}
            \g_o \\ \g_1 \\ \g_2
        \end{bmatrix}}_{\g(\q)} \\
        +
        & \underbrace{\begin{bmatrix}
                -\G_1 & -\G_2 \\ \J_1^\T & \zM \\ \zM & \J_2^\T
            \end{bmatrix}}_{\constraintmatrix^\T(\q)} \begin{bmatrix}
                \wrench_1 \\ \wrench_2
            \end{bmatrix}
            = \underbrace{\begin{bmatrix}
                \zM &\zM \\ \eye & \zM \\ \zM & \eye
            \end{bmatrix} }_{\inputmatrix}
            \begin{bmatrix}
                \jt_1 \\ \jt_2
            \end{bmatrix} \,.
    \end{split}
\end{align}
The matrices $\constraintmatrix$ and $\inputmatrix$ are used to represent the constraint and input matrices, respectively.
In the following, explicit dependencies on configuration and velocity variables are omitted for simplicity.

The system (\ref{eq:system}), including the pose constraints (\ref{eq:pose_constraints}), is a \gls{dae} with differentiation index~3 \cite{wanner1996solving}.
However, it is more convenient to enforce the constraints on the acceleration level using
\begin{equation}
    \constraintmatrix \dtwist = -\boldsymbol{\dot{\Lambda}}\twist = - \constraintvector  \,,
\end{equation}
with the constraint acceleration drift $\constraintvector$, instead of (\ref{eq:pose_constraints}) \cite{carpentier:hal-03271811}.

The dynamics of the coupled dual-arm system are thus given by
\begin{align}\label{eq:coupled_system}
    \begin{split}
        \M \dtwist + \constraintmatrix^\T \wrench &=\inputmatrix \jt -\n \\
         \constraintmatrix \dtwist &= -\constraintvector \,.
    \end{split}
\end{align}
The grasp matrix $\G(\q_o)$ does not have full column rank, resulting in a non-empty nullspace spanned by $\N(\q_o) \in \Rn{6\times12}$, which can be computed via singular value decomposition of $\G(\q_o)^\T$.
Any wrench in the nullspace, i.e., $\wrench_{\mathrm{int}} = \N(\q_o)\wrench$ for $\wrench \in \Rn{12}$, is an internal wrench that produces no object motion, satisfying $\G(\q_o)\N(\q_o)^\T \wrench_{\mathrm{int}}=\zV$ for all $\wrench_{\mathrm{int}} \in \Rn{6}$.

To avoid recomputation for varying object poses $\q_o$, we precompute the nullspace $\N_o$ in the local object frame using $\G_o$, then transform it to the inertial frame as needed.
The wrench transformation to the object frame is given by $\wrench_o = \hat{\Rotm}^o_I\wrench$, where $\hat{\Rotm}^o_I = \eye_4 \otimes \Rotm^o_I$.
This approach reduces online computational cost and avoids discontinuity issues due to nullspace non-uniqueness.

%% file: 01_content/problem/problem_formulation.tex
\input{01_content/problem/topp/topp.tex}

\input{01_content/problem/hqp/hqp.tex}

%% file: 01_content/problem/topp/topp.tex
\section{Time-Optimal Path Parameterization}\label{sec:topp}

This work builds on the decoupled motion planning approach and augments the \gls{topp3-ra} method from \cite{Dio2025} to handle redundantly actuated systems by treating internal wrenches as optimization variables.
We first derive the path-dependent kinematics and dynamics of the system, which are then used to express the constraints in a unified manner.
Finally, the \gls{topp3-ra} method is adapted to solve the resulting problem efficiently.

It is assumed that a geometric path for the object $\Transform_o(\s)$ and its corresponding path derivatives $\tilde{\twist}_o(\s)$, $\tilde{\twist}'_o(\s)$, $\tilde{\twist}''_o(\s)$ are provided with the path parameter $\s\in[0,1]$.
An analytical \gls{ik} algorithm is assumed to be available to compute a consistent joint path $\q_i(\s)$ satisfying pose constraints \eqref{eq:pose_constraints} for all $s\in[0,1]$.
For kinematically redundant manipulators, nullspace movement is parametrized by $\vm{\theta}_i(\s)$, yielding $\q_i(\s) = \q_{\mathrm{IK}}(\Transform_{\mathrm{o,i}}(\s), \vm{\theta}_i(\s))$.
The path derivatives up to third order must match, i.e.
\begin{equation}
	\frac{\diff^k \tilde{\twist}_{\mathrm{EE,i}}(\q_i(\s))}{\diff \s^k} = \frac{\diff^k \tilde{\twist}_{i,o}(\s)}{\diff \s^k}\,, \quad k=0,1,2 \quad \forall s\in[0,1]\,.
\end{equation}
Pose and joint derivatives are computed via central differences.
This results in path configurations $\q(\s)$ and consistent path derivatives $\tilde{\twist}(\s), \tilde{\twist}'(\s), \tilde{\twist}''(\s)$ for all $s\in[0,1]$.

\input{01_content/problem/topp/kin_dyn.tex}

\subsection{Unified Constraint Structure}
In the following a unified constraint structure is proposed. The bounds $\barbelow{\vm{\gamma}}_i$,$\vm{\bar{\gamma}}_i$ depend on context and can represent different limits like velocities or torques.
First-order constraints, i.e. Cartesian or joint velocity limits can be represented by
\[
	\barbelow{\vm{\gamma}}_1 \leq \vm{a}_1 \ds \leq \vm{\bar{\gamma}}_1 \,.
\]
The constraints can be equivalently expressed as
\begin{equation} \label{eq:vel_constraint2}
	0 \leq \dot{s}^2 \leq
	\min_{i}\left(\max\left(\frac{[\barbelow{\vm{\gamma}}_{1}]_i}{ [\vm{a}_1]_i},\frac{[\vm{\bar{\gamma}}_{1}]_i} {[\vm{a}_1]_i}\right)\right)^2 \,,
\end{equation}
assuming that the elements of the vector $[\barbelow{\vm{\gamma}}_{1}]_i\leq 0 \, \forall \, i $ and $[\vm{\bar{\gamma}}_{1}]_i \geq 0 \, \forall \, i$.
Second-order constraints can be represented by
\begin{equation}
	\barbelow{\vm{\gamma}}_2 \leq \vm{a}_2 \dds + \vm{b}_2 \ds^2 + \vm{F}_2  \wrench\internal  \leq \vm{\bar{\gamma}}_2 \,,
\end{equation}
while third-order constraints are given by
\begin{equation}
	\barbelow{\vm{\gamma}}_3 \leq ~ \ds(\vm{a}_3 \dds' + \vm{b}_3 \dds + \vm{c}_3 \ds^2 +
	 \vm{F}_3  \wrench\internal + \vm{d}_3)  \leq \vm{\bar{\gamma}}_3 \,.
\end{equation}
The vectors $\vm{a}_i$, $\vm{b}_i$, $\vm{c}_i$, $\vm{d}_i$, $\vm{F}_i$ correspond to the path-dependent coefficients in the equations \eqref{eq:general_kinematics}-\eqref{eq:tau_path_wrench}.

Combining all $c$ constraints in a proper manner, the admissible set is then given by
\begin{equation}\label{eq: Omega3}
	\Omega(\s,\ds,\dds,\dds',\wrench\internal) = \left\{
	\begin{bmatrix}\xPath \\ \uPath
	\end{bmatrix}
	\, \bigg| \,
	\vm{A}(\s)
	\begin{bmatrix}\xPath \\ \uPath
	\end{bmatrix} \leq
	\vm{b}(\s) + \frac{\vm{c}(\s)}{\ds} \right\}\,,
\end{equation}
with $\xPath^\T = \begin{bmatrix}
		\ds^2 & \dds
	\end{bmatrix} \in \Rn{2}$,
	$\uPath^\T = \begin{bmatrix}
		\dds' & \wrench\internal^\T
	\end{bmatrix} \in \Rn{m}$
	and $\vm{A}(\s) = \begin{bmatrix}
		\vm{A}_{\xPath}(\s) & \vm{A}_{\uPath}(\s)
	\end{bmatrix} \in \Rn{c\times(2+m)}$, assuming $\ds>0$.

For first- and second-order constraints, the elements in $\vm{c}(\s)$ are zero resulting in a linear set with respect to $\xPath$ and $\uPath$.
For third-order constraints, the elements in $\vm{c}(\s)$ are generally positive and non-zero. With $x_1 = \ds^2$, $[\vm{c}(\s)]_i/\ds$ is convex for all $i$ and the set becomes concave in $\xPath$ and $\uPath$.

\input{01_content/problem/topp/topp3-ra.tex}

%% file: 01_content/problem/topp/kin_dyn.tex
\subsection{Kinematics and Dynamics as a Function of the Path}

To compute the trajectory with the proposed method, the kinematics and dynamics of the system have to be expressed as a function of the path parameter $\s$.
The system velocity, acceleration, and jerk are computed by
\begin{subequations} \label{eq:general_kinematics}
    \begin{align}
        \begin{split}
            \twist & = \tilde{\twist}\ds \,,
        \end{split}               \\
        \begin{split}
            \dtwist & = \tilde{\twist}\dds + \tilde{\twist}'\ds^2	\,,
        \end{split} \\
        \begin{split}
            \ddtwist & = \ds\left( \tilde{\twist}\dds' + 3\tilde{\twist}'\dds + \tilde{\twist}''\ds^2 \right) \,,
        \end{split}
    \end{align}
\end{subequations}
with $\ddds= \ds\dds'$ \cite{Debrouwere2013}.

Inserting the kinematics \eqref{eq:general_kinematics} into the coupled object dynamics \eqref{eq:obj_dyn_augmented} and manipulator dynamics \eqref{eq:man_dyn_augmented}
results in the path-dependent dynamics
\begin{equation} \label{eq:obj_dyn_augmented_twist}
    \underbrace{\M_o\tilde{\twist}_o}_{\wrench_m}\dds + \underbrace{\left(\M_o\tilde{\twist}_o' + \n_o\right)}_{\wrench_c}\ds^2 + \underbrace{\g_o}_{\wrench_g} - \G\wrench = \zV
\end{equation}
and
\begin{equation} \label{eq:tau_path}
    \jt_i = \underbrace{\M_i\q'_i}_{\jt_{i,m}}\dds+\underbrace{(\n_i+\M_i\q_i'')}_{\jt_{i,c}} \ds^2  + \underbrace{\vm{g}_i}_{\jt_{i,g}} + \J_i^\T\wrench_i \,.
\end{equation}
As noted before, the grasp matrix $\G$ lacks full column rank. Therefore, $\G$ is not invertible and there exists no unique solution for $\wrench$. Because a tight grasp is assumed, there is no need to limit $\wrench$ directly.
Here, only an internal wrench $\wrench_{\internal}$ should be constrained avoiding the destruction of the object.
The object wrench $\wrench$ can be decomposed using the pseudo-inverse $\G\pinv$ and the nullspace matrix $\vm{\mathcal{N}}$ of the grasp matrix $\G$.
With the internal wrench $\wrench_{\internal}$, it is given by
\begin{equation} \label{eq:wrench_object_path}
    \wrench(\s,\ds,\dds,\wrench_{\internal}) = \vm{\mathcal{N}}^\T\wrench_{\internal} + \G\pinv( \wrench_m\dds + \wrench_c\ds^2 + \wrench_g)\,.
\end{equation}

The insertion of \eqref{eq:wrench_object_path} in \eqref{eq:tau_path} for both arms results in the path dynamics and the internal wrench dependent joint torques
\begin{equation} \label{eq:tau_path_wrench}
    \begin{aligned}
        \jt & = \underbrace{\left(\jt_{m} + \J^\T \G\pinv \wrench_m \right)}_{\hatjt_{m}}\dds +
        \underbrace{\left(\jt_{c} + \J^\T \G\pinv \wrench_c \right)}_{\hatjt_{c}}\ds^2 \\
        & \quad+ \underbrace{\jt_{g} + \J^\T \G\pinv \wrench_g }_{\hatjt_{g}}  + \underbrace{\J^\T\vm{\mathcal{N}}^\T}_{\hatJ^\T}\wrench_{\internal} \,.
    \end{aligned}
\end{equation}
This formulation reduces the number of optimization variables in the \gls{topp} problem and
differs from \cite{Dio2024} where long computation times due to the high number of optimization variables were observed.
However, this approach is not suitable for friction constraints where the interaction wrenches themselves are constrained, e.g. have to stay in a friction cone.

%% file: 01_content/problem/topp/topp3-ra.tex
\subsection{Augmented TOPP3-RA Method} \label{sec:topp3_ra_method}

The resulting \gls{topp} problem minimizes the cycle time $T\cycle=\int_{0}^{T\cycle}{1~ \diff t}$ subject to boundary conditions, the admissible set $\fSet(\s,\ds,\dds,\dds',\wrench\internal)$, and the path dynamics.
To solve this efficiently, the \gls{topp3-ra} method from \cite{Dio2025} is augmented to treat internal wrenches $\wrench\internal$ as additional optimization variables.
This extends the approach of \cite{PhamRA2017} to accommodate third-order constraints.

The \gls{topp} problem is reformulated as an \gls{ocp}
\begin{mini}|s|
    {\ds(\s)}{ T\cycle = \int_0^1\frac{1}{\ds}~\diff \s}
    {\label{eq:TOPP3_path}}{}
    \addConstraint{\ds(0) = \ds_0 \,,\quad \dds(0) = \dds_0}
    \addConstraint{\ds(1) = \ds_T \,,\quad \dds(1) = \dds_T}
    \addConstraint{(\ds^2)' = 2\dds \,, \quad (\dds)' = \dds'}
    \addConstraint{\left(\s,\ds,\dds,\dds',\wrench\internal\right) \in \as(\s,\ds,\dds,\dds',\wrench\internal)}
    \addConstraint{\forall \s \in [0,1]}
\end{mini}
in the path domain which is then discretized with $N+1$ grid points $\s_0, \s_1, \dots, \s_N$ with $\s_0 = 0$ and $\s_N = 1$.
This results in the discrete linear path dynamics
\begin{equation}\label{eq: discrete_dynamics}
    \begin{aligned}
        \underbrace{\begin{bmatrix}\ds^2_{k+1} \\ \dds_{k+1} \end{bmatrix}}_{\boldsymbol{x}_{k+1}} = \underbrace{\begin{bmatrix}1 & 2\Delta \\ 0 & 1 \end{bmatrix}}_{\boldsymbol{\Phi}_k} \underbrace{\begin{bmatrix}\ds^2_{k} \\ \dds_{k} \end{bmatrix}}_{\boldsymbol{x}_{k}} + \underbrace{\begin{bmatrix} \Delta^2 \\ \Delta \end{bmatrix}}_{\boldsymbol{\Gamma}_k} \dds'_{k}\,,
    \end{aligned}
\end{equation}
with path step $\Delta = \s_{k+1} - \s_k$.
The admissible set $\as_k = \as(\s_k,\ds_k,\dds_k,\dds'_k,\wrench_{\mathrm{int},k})$ evaluated at $\s_k$ can be expressed as linear inequalities in state and input, except for nonconvex third-order constraint terms. These are convexified around a reference velocity profile, initially computed without third-order constraints. The solution is then used for the next iteration and the process is repeated sequentially until convergence. Typically, three iterations are sufficient.

Each iteration of the algorithm employs a two-pass scheme.
Starting with a backward pass, it recursively computes controllable sets $\cSet_k$ from the terminal state $\boldsymbol{x}_{N}$ to the initial state $\boldsymbol{x}_{0}$ as
\begin{equation}
    \cSet_k = \left\{\boldsymbol{x}_k \in \Rn{2} ~|~ \Pi_x \left( \Omega_{k} \bigcap \cSet_{k|k+1} \right)\right\} \,,
\end{equation}
where
\begin{equation}
    \cSet_{k|k+1} = \{\boldsymbol{x}_k \in \Rn{2}\,,  \dds'_{k} \in \Rn{1} \,|\, (\boldsymbol{\Phi}\boldsymbol{x}_k + \boldsymbol{\Gamma}\dds'_{k}) \in \cSet_{k+1}\}
\end{equation}
is the backward-propagated controllable set.
The operator $\Pi_x:\Rn{2+m}\rightarrow\Rn{2}$ projects from the state-input space to the state space.
In the forward pass, the optimal trajectory is obtained by maximizing $\dds'_{k}$ subject to $\Omega_k$ and $\cSet_{k|k+1}$, starting from $\boldsymbol{x}_{0}$.

This approach achieves linear time complexity in the number of discretization steps $N$.
However, the backward propagation of controllable sets can become computationally expensive due to constraint growth.
To mitigate this, \cite{Dio2025} proposes a \gls{vmrs} algorithm for under-approximating the controllable sets.
Because no viscous friction is considered and the admissible set is convexified, the ray shooting can be performed by solving linear programs in this work.

Compared to \cite{Dio2025}, this work extends the \gls{topp3-ra} method by incorporating internal wrenches $\wrench\internal$ as additional optimization variables in the admissible set $\as(\s,\ds,\dds,\dds',\wrench\internal)$.
In contrast to \cite{Dio2024}, internal wrenches are directly optimized rather than wrenches at contact points which reduces the number of optimization variables but restricts applicability to tight grasps.
While these internal wrenches do not alter the path dynamics, they enable higher path accelerations when considering torque limits.
The resulting solution is generally non-unique, as multiple internal wrench distributions can produce identical system accelerations.
This redundancy is not resolved within the \gls{topp} framework. Instead, it is managed by the subsequent \gls{hqp} controller, which selects an optimal internal wrench distribution during trajectory tracking.

%% file: 01_content/problem/hqp/hqp.tex
\section{Hierarchical Quadratic Programming Control}\label{sec:hqp}

With the dynamic model and the reference trajectory in place, the control architecture for the cooperative dual-arm object control is presented. It is based on the \gls{hqp} control of \cite{Dio2023} but requires modifications to incorporate the time-optimal reference trajectory.

\input{01_content/problem/hqp/problem_definition.tex}

\input{01_content/problem/hqp/hqp_formulation.tex}

%% file: 01_content/problem/hqp/problem_definition.tex
Given a reference trajectory $(\q_o\reference,\twist_o\reference,\dtwist_o\reference)$ for the object and $(\q_i\reference,\dq{i}\reference,\ddq{i}\reference)$ for the manipulators, a virtual system acceleration $\dtwist\virtual$ is calculated with the measured system states $(\q_o,\q_1,\q_2,\twist_o,\dq{1},\dq{2})$.

\subsection{Virtual Object Control}

For the object, the virtual acceleration is computed by
\begin{equation} \label{eq:virt_obj_acc}
	\dtwist_o\virtual= \dtwist_o\reference + \vm{K}_{d,o} (\twist_o\reference-\twist_o) + \vm{h}_{\vm{K}_{p,o}}( \q_o\reference, \q_o),
\end{equation}
where the dampening matrix $\vm{K}_{d,o}$ and the geometrically consistent stiffness $\vm{h}_{\vm{K}_{p,o}}$ \cite{Caccavale2008} are used to decouple the translational and rotational movement and provide a desired closed-loop behavior.

Differently from the original approach in \cite{Dio2023} which was based on \cite{Erhart2016}, the virtual joint accelerations are not computed by a kinematic coordination but directly from
\begin{equation} \label{eq:virt_joint_acc}
	\ddq{i}\virtual= \ddq{i}\reference + \vm{K}_{d,i} (\dq{i}\reference-\dq{i}) + \vm{K}_{p,i}(\q_i\reference-\q_i),~i=1,2,
\end{equation}
as the kinematically consistent reference trajectory for the joints is available from the \gls{topp} solution.
The virtual joint acceleration for the manipulators is calculated with the dampening matrix $\vm{K}_{d,i}$ and the proportional gain matrix $\vm{K}_{p,i}$.

The resulting virtual system acceleration
\[
\dtwist\virtual = \begin{bmatrix} (\dtwist_o\virtual)^\T & (\ddq{1}\virtual)^\T & (\ddq{2}\virtual)^\T \end{bmatrix}^\T
\]
serves as the reference for the subsequent control level to compute motor torques $\jt$.
Since the virtual system acceleration may violate system constraints when the system deviates from the reference trajectory, the proposed method enforces constraint satisfaction through an optimization framework, detailed in \secref{sec: hqp_optimal inverse dynamics control}.

Previous approaches, such as those in \cite{Erhart2016}, computed end-effector accelerations and interaction wrenches separately without considering joint torque or internal wrench limits.
This decoupled strategy is problematic for heavy objects exceeding the manipulator's load capacity, as clipped joint torques can create inconsistencies between end-effector wrenches and accelerations.
The proposed coupled inverse dynamics control scheme addresses this limitation by explicitly enforcing system constraints and bounds through an optimal control formulation.

\subsection{Optimal Coupled Inverse Dynamics Control} \label{sec: hqp_optimal inverse dynamics control}

Consider the constrained system \eqref{eq:coupled_system}, for which an optimal control action shall be computed online.
Going forward, a static optimization problem is formulated as
\begin{equation} \label{eq:optimization_system_acc}
	\begin{aligned}
		\min_{\dtwist, \wrench, \jt} & \quad \frac{1}{2} \vnorm{\dtwist - \dtwist\virtual}{2}{\vm{W}_{\dtwist}}                                                \\
		\st                           & \quad \begin{bmatrix}\M & \constraintmatrix^\T & -\inputmatrix \\ \constraintmatrix & \zM & \zM \end{bmatrix} \begin{bmatrix}
			\dtwist \\ \wrench \\ \jt
		\end{bmatrix} = - \begin{bmatrix}
			\n + \g \\ \constraintvector
		\end{bmatrix},                                   \\
		                              & \quad \barbelow{\jt} \leq \jt \leq \bar{\jt}\,,                                                                            \\
		                              & \quad \barbelow{\wrench}_\mathrm{int} \leq \N_o \hat{\vm{R}}^o_I \wrench \leq \bar{\wrench}_\mathrm{int} \,,
	\end{aligned}
\end{equation}
which has to be solved in every time step.
The main goal is to minimize the error between a virtual system acceleration $\dtwist\virtual$ and a feasible system acceleration $\dtwist$ in the least squares sense given the system dynamics, system constraints and the joint torque limits.
Moreover, inequality constraints on the wrenches are considered.
These are projected onto the null space basis of the grasp matrix via $\N_o\in \Rn{6\times12}$.

The cost function of problem \eqref{eq:optimization_system_acc} is positive semi-definite. Therefore, the solution is not unique and given by a solution space which is constrained by the nullspace of the equality constraints.
While additional penalization of the joint torques or the interaction wrench with reduced weighting is possible to eliminate the redundancy, it results in a poorly conditioned optimization problem and requires additional tuning of the weights to reduce the tracking error even when exact tracking is attainable in the standard formulation.

Kinematically redundant single arm robots face a similar challenge when the main control objective is defined in Cartesian task space.
The combination of joint torques that results in the desired end effector acceleration is not unique and the space for optimal joint accelerations is spanned by the null space of the Jacobian \cite{Hoffman2018}. Therefore, additional objectives can be formulated w.r.t. the null space such as singularity or obstacle avoidance while defining the problem as a single~\gls{qp}. In this case, a decoupled solution can then be found which optimizes both objectives independently.

However, in the presence of inequality constraints, the desired end effector acceleration could become infeasible or the rank of the null space can reduce. Therefore, an explicit hierarchy is required when a constraint becomes active and the solution space shrinks. To incorporate these thoughts, the problem has to be reformulated as a lexicographic optimization problem, explicitly enforcing a hierarchical structure~\cite{Escande2014}.

%% file: 01_content/problem/hqp/hqp_formulation.tex
\subsection{HQP Formulation} \label{sec:HQP_fromulation}

The resulting minimizer $\dtwist^*$ is the optimal system acceleration considering the constraints. The solution space has at most 6 \glspl{dof} constrained on the nullspace of the equality constraints of \eqref{eq:optimization_system_acc}, because the cost function is positive semi-definite. Therefore, it is possible to impose additional objectives, such as minimizing joint torques or the internal wrench without affecting the optimal acceleration $\dtwist^*$.
Thus, a secondary, reduced optimization problem is formulated as
\begin{equation} \label{eq:optimization_TorqueWrench}
\begin{aligned}
    \min_{\wrench,\jt} & \quad \frac{1}{2} \vnorm{ \begin{bmatrix} \wrench \\ \jt \end{bmatrix}}{2}{\mathbf{W}_{\wrench,\jt}} \\
    \st & \quad\begin{bmatrix} \constraintmatrix^\T & -\inputmatrix \end{bmatrix} \begin{bmatrix}  \wrench \\ \jt \end{bmatrix} = -\n -\g - \M\dtwist^*\,,\\
    & \quad \barbelow{\jt} \leq \jt \leq \bar{\jt}\,, \\
    & \quad \barbelow{\wrench}_\mathrm{int} \leq \N_o \hat{\mathbf{R}}^o_I \wrench \leq \bar{\wrench}_\mathrm{int} \,.
\end{aligned}
\end{equation}

This \gls{qp} \eqref{eq:optimization_TorqueWrench} is solved after the first \gls{qp} \eqref{eq:optimization_system_acc} and explicitly includes the optimal system acceleration $\dtwist^{*}$ in the equality constraint, while the inequality constraints remain unchanged.
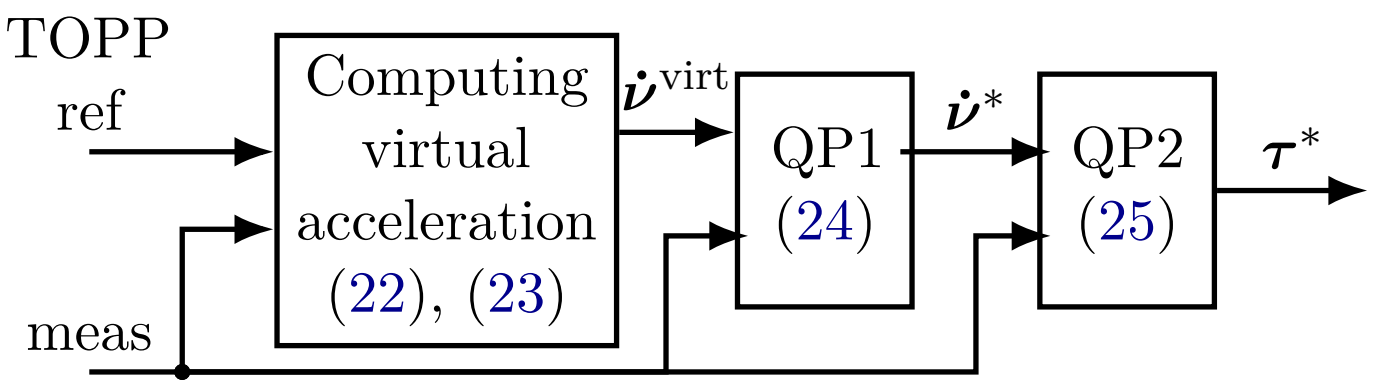
\begin{figure}[tb]
	\centering
	\input{02_figures/hqp_architecture/hqp_architecture.tex}
	\caption{\gls{hqp} control architecture for optimal dual-arm cooperative object manipulation with reference \textit{ref} and measurements \textit{meas}.}
	\label{fig:hqp_architecture}
\end{figure}
Although the \gls{hqp} formulation in \cite{Escande2014} is more efficient for many hierarchies, the problem at hand has only two hierarchies and does not benefit as much from this algorithm.
Therefore, the \gls{hqp} problem here is formulated as a cascaded structure because it is easier to implement and can leverage efficient solvers for standard \glspl{qp}.

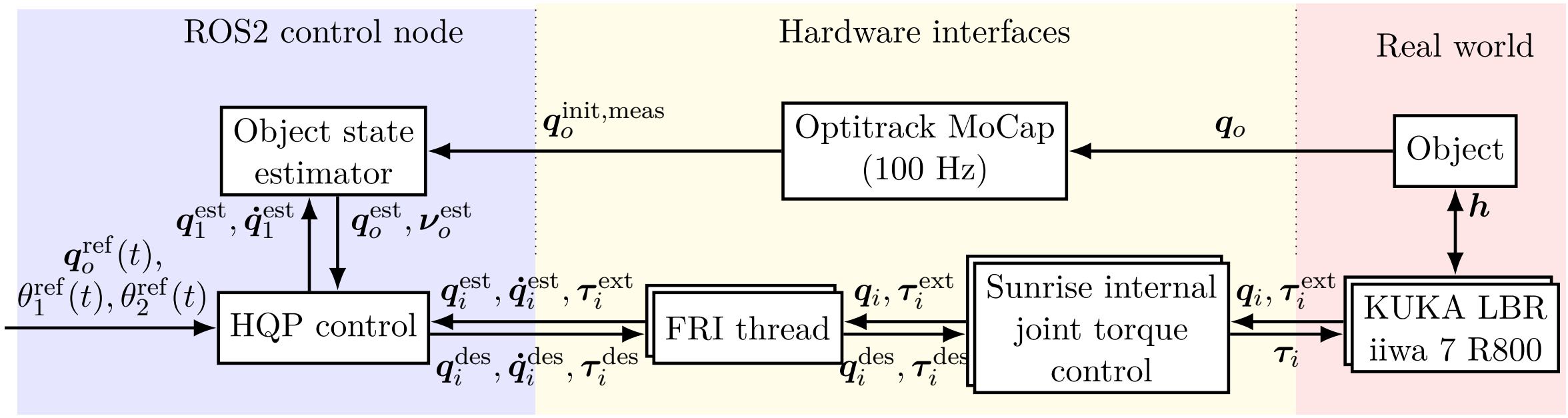
\begin{figure}[tb]
    \centering
    \resizebox{\textwidth}{!}{\input{02_figures/system_architecture/system_architecture.tex}}
    \caption{System architecture for experimental validation of the cooperative dual-arm system.}
    \label{fig:system_setup}
\end{figure}

The control architecture for the manipulators is broken down into three distinct blocks, as seen in \figref{fig:hqp_architecture}.
In the first block, the virtual system accelerations $\dtwist\virtual$ are calculated with \eqref{eq:virt_obj_acc} and \eqref{eq:virt_joint_acc}.
The last two blocks comprise the \gls{hqp} formulation described by the cascaded \gls{qp} structure.
In the primary \gls{qp} the virtual acceleration $\dtwist\virtual$ is tracked at best in the least square sense via \eqref{eq:optimization_system_acc}.
This results in an optimal system acceleration $\dtwist^*$.
The secondary \gls{qp} \eqref{eq:optimization_TorqueWrench} eliminates the redundancy by optimizing joint torques and wrenches on the solution space of the first~\gls{qp}.
The optimal interaction wrenches $\wrench^*$ which compensate the dynamics of the object are implicitly incorporated in the joint torques.
Therefore, the optimal joint torque $\jt^*$ can be directly applied to the manipulators.

%% file: 02_figures/hqp_architecture/hqp_architecture.tex
\def\boxshift{0.666em}

\begin{tikzpicture}
    [box/.style={draw, fill=white, minimum height=3em, minimum width=3em, line width = \plotlinewidth, align=center},
    link/.style={-Latex, line width = \plotlinewidth},
    arr/.style={{Circle[length=3pt]}-Latex, shorten <=-1.5pt, line width = \plotlinewidth},
    clip = true]

\node(in_1) [minimum height=2em] at (0,0) {};
\node(in_2) [minimum height=2em] at (0,-1.2) {};

\node [box, minimum height=4em] (Null) at (2.5,0) {\Umbruch{Computing \\ virtual \\acceleration\\ \eqref{eq:virt_obj_acc}, \eqref{eq:virt_joint_acc}}};
\node [box, minimum height=4em, yshift=0em] (QP_1) at (5,0) {\Umbruch{QP1\\ \eqref{eq:optimization_system_acc}}};
\node [box, minimum height=4em, yshift=0em] (QP_2) at (7,0) {\Umbruch{QP2\\ \eqref{eq:optimization_TorqueWrench}}};

\draw [link] ([yshift=\boxshift] in_1.east) node[above]{\Umbruch{TOPP\\ref}} -- ([yshift=\boxshift] Null.west);

\draw [link] (in_2.east) node[above]{meas} -- (0.75,-1.2)  -- (0.75,-\boxshift) -- ([yshift=-\boxshift] Null.west);

\draw [arr] (0.75,-1.2) |- ++ (5.25,0) |- (6.5,-0.3);

\draw [arr] (0.75,-1.2) |- ++ (3.2,0) |- (4.5,-0.3);

\draw[link] ([yshift=1em] Null.east) -- ([yshift=1em] QP_1.west) node[midway,above]{$\dtwist\virtual$};

\draw[link] (5.5,\boxshift)  -- (6.5,\boxshift) node[midway,above]{$\dtwist^*$};

\draw[link] (QP_2.east) -- ++ (1,0) node[midway,above]{$\jt^*$};

\end{tikzpicture}

%% file: 02_figures/system_architecture/system_architecture.tex
\def\gs{23pt}
\def\hone{0\gs}
\def\htwo{-1.5*\gs}

\tikzstyle{block} = [rectangle, draw, text centered, minimum height=1*\gs, fill=white, line width = \plotlinewidth]
\tikzstyle{multiblock} = [rectangle, draw, text centered, minimum height=2.5em, fill=white, line width = \plotlinewidth]
\tikzstyle{arrow} = [line width = \plotlinewidth,-Latex,>=stealth]
\tikzstyle{arrowinv} = [line width = \plotlinewidth,Latex-,>=stealth]
\tikzstyle{doublearrow} = [line width = \plotlinewidth,Latex-Latex,>=stealth]

\begin{tikzpicture}[node distance=2cm]

    \fill[blue!10] (-3.5,-2.2) rectangle (2.5,3*\gs);
    \fill[yellow!10] (3*\gs,-2.2) rectangle (13.75*\gs,2.5);
    \fill[red!10] (13.75*\gs,-2.2) rectangle (14.2,2.5);

    \node (node) at (0*\gs, \hone+1.7*\gs) {ROS2 control node};
    \node (estimator) at (0*\gs, \hone) [block] {\Umbruch{Object state estimator}};
    \node (hqp) at (0*\gs, \htwo) [block] {HQP control};

    \draw [dotted] (3*\gs, \hone+2*\gs) -- (3*\gs, -0.5*\gs);

    \node (node) at (8.5*\gs, \hone+1.7*\gs) {Hardware interfaces};
    \node (optitrack) at (8.5*\gs, \hone) [block] {\Umbruch{Optitrack MoCap (100 Hz)}};

    \node (internal) at (11*\gs-0.1*\gs, \htwo +0.1*\gs) [block] {\Umbruch{Sunrise internal joint-torque control}};
    \node (internal) at (11*\gs, \htwo) [block] {\Umbruch{Sunrise internal joint torque control}};
    \node (fri) at (6*\gs-0.1*\gs, \htwo +0.1*\gs) [block] {FRI thread};
    \node (fri) at (6*\gs, \htwo) [block] {FRI thread};

    \draw [dotted] (13.75*\gs, \hone+2*\gs) -- (13.75*\gs, -0.5*\gs);

    \node (node) at (16*\gs, \hone+1.5*\gs) {\Umbruch{Real world}};

    \node (object) at (16*\gs, \hone) [block] {Object};
    \node (kuka) at (16*\gs-0.1*\gs, \htwo+0.1*\gs) [multiblock] {\Umbruch{KUKA LBR \\ iiwa 7 R800}};
    \node (kuka) at (16*\gs, \htwo) [multiblock] {\Umbruch{KUKA LBR  \\ iiwa 7 R800}};

	\draw [arrow] (object) -- node[midway, above] {$ \q_o$} (optitrack);
    \draw [arrow] (optitrack) -- node[midway, above] {$ \q_o^\mathrm{init, meas}$} (estimator);

	\draw [arrowinv] ([xshift= 0.2*\gs]  hqp.north) -- ++(0,\hone)  -| node[midway, right] {$ \q_o^\mathrm{est},  \twist_o^\mathrm{est}$} ([xshift= 0.2*\gs] estimator.south);
    \draw [arrow] ([xshift= -0.2*\gs]  hqp.north) -- ++(0,\hone)  -| node[midway, left] {$ \q_1^\mathrm{est},   \dq{1}^\mathrm{est}$} ([xshift= -0.2*\gs] estimator.south);

	\draw [arrow] ([xshift= -0.1*\gs,yshift=+2.0\gs] kuka.west) -- node[midway, above] {$ \q_i,  \jt_i^\mathrm{ext}$} ([yshift=+2.0\gs] internal.east);
    \draw [arrowinv] ([xshift= -0.1*\gs,yshift=-2.0\gs] kuka.west) -- node[midway, below] {$ \jt_i$} ([yshift=-2.0\gs] internal.east);

    \draw [arrow] ([xshift= -0.1*\gs,yshift=+2.0\gs] internal.west) -- node[midway, above] {$ \q_i,  \jt_i^\mathrm{ext}$}  ([ yshift=+2.0\gs] fri.east);
    \draw [arrowinv] ([xshift= -0.1*\gs,yshift=-2.0\gs] internal.west) -- node[midway, below] {$ \q_i^\mathrm{des},  \jt_i^\mathrm{des}$} ([yshift=-2.0\gs] fri.east);

    \draw [arrow] ([xshift= -0.1*\gs,yshift=+2.0\gs] fri.west) -- node[midway, above] {$ \q_i^\mathrm{est},   \dq{i}^\mathrm{est},  \jt_i^\mathrm{ext}$}  ([yshift=+2.0\gs] hqp.east);
    \draw [arrowinv] ([xshift= -0.1*\gs,yshift=-2.0\gs] fri.west) -- node[midway, below] {$ \q_i^\mathrm{des},   \dq{i}^\mathrm{des},  \jt_i^\mathrm{des}$} ([yshift=-2.0\gs] hqp.east);

	\draw [doublearrow] ([yshift= 0.1*\gs,] kuka.north) -- ++(0,\hone) -| node[midway, right] {$\boldsymbol{h}$} (object.south);

    \draw [arrow] ([xshift=-3*\gs] hqp.west) -- node[midway, above] {\Umbruch{$ \q_o^\mathrm{ref}(t)$, $\theta_1^\mathrm{ref}(t), \theta_2^\mathrm{ref}(t)$}} (hqp.west);

\end{tikzpicture}

%% file: 01_content/experimental_evaluation.tex
\section{Experimental Evaluation} \label{sec:experimental_evaluation}

This section evaluates the time-optimal manipulation of heavy objects using cooperative dual-arm robots. First, the hardware and software setup are described, then results from the offline trajectory planning and the experiments on the real system are presented.

\input{01_content/experimental_and_software_setup.tex}

\input{01_content/scenario_B.tex}

%% file: 01_content/experimental_and_software_setup.tex
\subsection{Experimental and Software Setup}

\begin{table}[tb]
    \caption{Joint limits used in the experiments.}
    \label{tab:experimental_limits}
    \centering
    \begin{tabular}{lccccccc}
        \toprule
        Joint & $\bar{\dq{}}$ [\si{\degree\per\second}] & $\bar{\ddq{}}$ [\si{\degree\per\second\squared}]& $\bar{\q}^{(3)}$ [\si{\degree\per\second\cubed}] & $\bar{\jt}$ [\si{\newton\meter}] \\
        \midrule
        $A_1$ & 78.4 & 292 & 7976 & 166 \\
        $A_2$ & 78.4 & 292 & 6188 & 166 \\
        $A_3$ & 80.0 & 299 & 7976 & 100 \\
        $A_4$ & 104.0 & 388 & 7746 & 40 \\
        $A_5$ & 112.0 & 419 & 15493 & 100 \\
        $A_6$ & 144.0 & 540 & 25669 & 30 \\
        $A_7$ & 144.0 & 540 & 25669 & 30 \\
        \bottomrule
    \end{tabular}
\end{table}

The system consists of two \textit{KUKA LBR iiwa 7 R800} manipulators (\SI{7}{\kilogram} nominal payload each) mounted at calibrated poses.
The grasp frames are predetermined and fixed during execution.
Each arm is equipped with joint torque sensors and \textit{OnRobot 2FG7} grippers (\SI{1.2}{\kilogram}, \SI{140}{\newton} grasp force).
All experiments are conducted with a load of \SI{16}{\kilo\gram}, exceeding the nominal payload of both arms.

The system architecture is illustrated in \figref{fig:system_setup}.
Communication with the robots is done via a \gls{fri} thread at \SI{1}{\kilo\hertz} using the Sunrise internal torque controller.
An \textit{Optitrack} \gls{mocap} system initializes the object pose but due to a limited update rate, it is not used for closed-loop control.
Instead, the pose and twist are estimated from forward kinematics with filtered signals.

The \gls{hqp} controller runs on an external computer (Intel Core i5-9400, \SI{2.90}{\giga\hertz}) as a ROS2 node in C++, using \textit{Eigen} for linear algebra computations, \textit{qpOASES} for solving \gls{qp} problems, and \textit{Pinocchio} for evaluating robot dynamics.
The \gls{topp} solution is computed offline in Python.
Reference poses $\Transform_{\mathrm{o}}\reference(t)$ and joint positions $\q_i\reference(t)$ are stored. Velocities and accelerations are computed via central differences at runtime.

Nominal system limitations for joint velocities $\bar{\dq{}}$ and torques $\bar{\jt}$ are provided by the manufacturer while acceleration $\bar{\ddq{}}$ and jerk $\bar{\q}^{(3)}$ limits are estimated based on experimental observations.
During the experiments, more conservative limits are applied, as detailed in \tabref{tab:experimental_limits}.
For the velocity and jerk limits, $80\%$ of the nominal values are used, while torque limits are reduced by \SI{10}{\newton\meter} for all joints and \SI{40}{\newton\meter} for joint $A_4$ of both robots.
The internal wrench limits are set to \SI{30}{\newton} for forces and \SI{30}{\newton\meter} for moments in all directions.
This allows for safety margins and still illustrates that the robot works beyond its nominal payload when cooperating.
Acceleration limits are set to $60\%$ of the estimated values as falsely estimated object properties like inertias can lead to larger tracking errors when following aggressive trajectories.

%% file: 01_content/scenario_B.tex
\subsection{Planning Results} \label{sec: planning_results}

\begin{figure}[tb]
	\centering
	\begin{subfigure}[t]{0.59\textwidth}
		\centering
		\includegraphics[width=\linewidth]{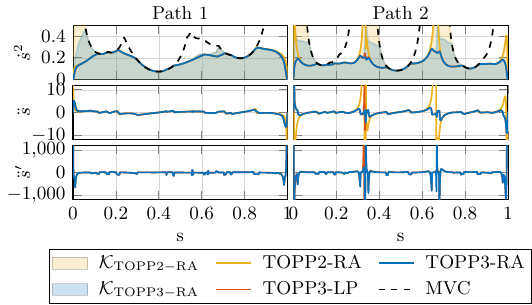}
		\caption{Results of $\ds^2$, $\dds$, and $\dds'$ for different \gls{topp} methods and paths. Shaded areas represent controllable sets of the \gls{topp-ra} methods.}
		\label{fig:planning_comparison}
	\end{subfigure}

	\medskip
	\begin{subfigure}[t]{0.59\textwidth}
		\centering
		\includegraphics[width=\linewidth]{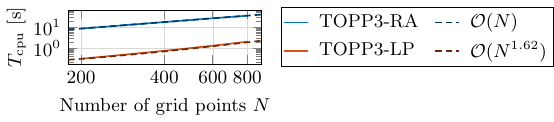}
		\caption{Mean computation time per iteration for different \gls{topp} methods and grid points along with complexity estimates.}
		\label{fig:planning_comparison2}
	\end{subfigure}
	\caption{Planning results for different paths and methods.}
	\label{fig:planning_results}
\end{figure}

The reference trajectory is obtained using the proposed method of \secref{sec:topp3_ra_method} with $N=500$ grid points and $k=20$ ray expansions per backward step, solved via \texttt{cvxpy}/\texttt{GLPK} \cite{diamond2016cvxpy}.
Increasing $k$ did not significantly improve the results.
For comparison, TOPP3-LP \cite{Dio2024} uses the adapted constraint structure of this work, solving a large sparse \gls{lp} via \texttt{scipy}'s \texttt{linprog}/\texttt{highs}.
Both methods perform three iterations linearizing the concave jerk constraints around the previous solution.
Both are initialized with a jerk-unconstrained TOPP2-RA \cite{PhamRA2017} solution.

A comparison of the resulting time-optimal solutions for two exemplary paths is shown in \figref{fig:planning_comparison} showing the \gls{mvc}.
The first path avoids singularities, while the second path includes path singularities where the path derivatives are zero.
The proposed \gls{topp-ra} method with jerk limits achieves results comparable to the collocation-based approach from \cite{Dio2024}.
At singular points, minor deviations occur, but the proposed method can reduce the velocity spike compared to the collocation-based approach, resulting in smoother trajectories.
Note that the internal wrench is not visualized as in the current stage its redundancy is not resolved and shows significant jitter over the path. Although a postprocessing step could be applied as suggested in \cite{Dio2024}, the controller in \secref{sec:hqp} directly handles the redundancy, making this step unnecessary.

\begin{figure}[tb]
	\centering
	\includegraphics[width=0.55\textwidth]{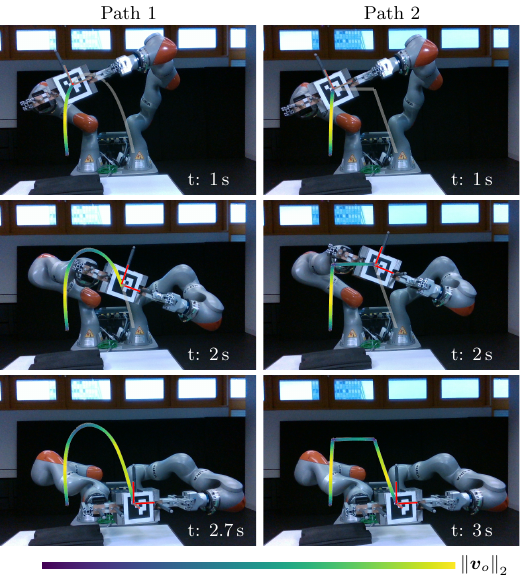}
	\caption{Video frames at different time steps. The video is provided as an ancillary file of this preprint.}
	\label{fig:real_01_video_frames}
\end{figure}

A direct comparison of computation times is not meaningful, as \gls{topp3-ra} was implemented in Python, whereas TOPP3-LP leverages the highly optimized \texttt{highs} solver written in C++.
Figure \ref{fig:planning_comparison2} shows the computation time per iteration for both \gls{topp3-ra} and TOPP3-LP as a function of grid points.
The complexity analysis shows that \gls{topp3-ra} scales linearly with the number of grid points $N$, while TOPP3-LP exhibits super-linear scaling with approximately $\mathcal{O}(N^{1.62})$.
This difference arises because \gls{topp3-ra} solves multiple small \glspl{lp}, whereas TOPP3-LP solves a single large sparse \gls{lp}.
For reference, the efficiently implemented \gls{topp-ra} method in \cite{PhamRA2017} solves a related problem without jerk constraints in a few milliseconds, using 2 \glspl{lp} with 2 variables per backward step and 1 \gls{lp} with 1 variable per forward step.
The proposed \gls{topp3-ra} method extends this by solving $k$ \glspl{lp} with 3 variables in the backward pass and one \gls{lp} with 1 variable per forward step.
These observations suggest that with an efficient C++ implementation, the proposed method could outperform the collocation-based approach for problems with jerk constraints at fine discretizations.

\subsection{Experimental Results} \label{sec: experimental_results}

\begin{figure}[tb]
	\centering
	\includegraphics[width=0.59\textwidth]{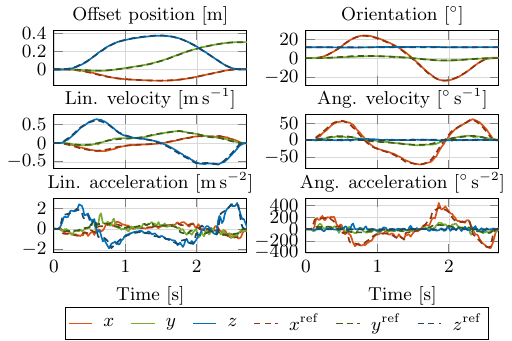}
	\caption{Object trajectory for path 1. The position is offset by the initial position of the object. The orientation is represented in roll-pitch-yaw angles.}
	\label{fig:real_01_object_trajectory}
\end{figure}

\begin{figure}[tb]
	\centering
	\begin{subfigure}[t]{0.59\textwidth}
		\centering
		\includegraphics[width=\linewidth]{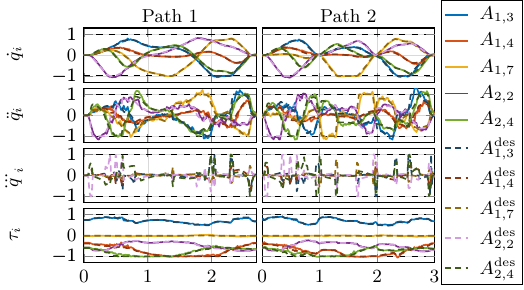}
		\caption{Selected joint trajectories and torques of axis $A_{i,j}$ for robot $i$ and joint $j$, normalized to maximum value.}
		\label{fig:real_01_joint_trajectory}
	\end{subfigure}

	\medskip
	\begin{subfigure}[t]{0.485\textwidth}
		\centering
		\includegraphics[width=\linewidth]{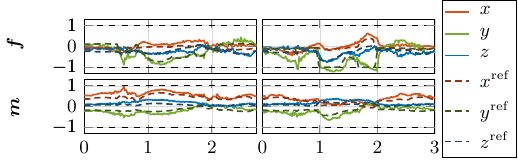}
		\caption{Normalized internal wrench in object frame.}
		\label{fig:real_01_internal_wrench}
	\end{subfigure}
	\hfill
	\begin{subfigure}[t]{0.485\textwidth}
		\centering
		\includegraphics[width=\linewidth]{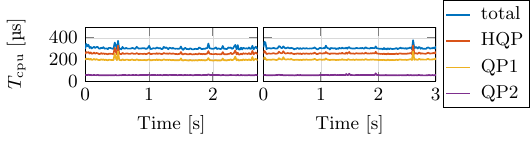}
		\caption{Computation time of the \gls{hqp} controller and its components.}
		\label{fig:real_01_cpu_time}
	\end{subfigure}
	\caption{Experimental results for path 1 and 2.}
	\label{fig:real_01_experimental_results}
\end{figure}

Figure \ref{fig:real_01_video_frames} shows frames from the video of the experimental evaluation for path 1 and 2.
The video of the experiment is provided as an ancillary file of this preprint.
The object trajectory for path 1 is depicted in \figref{fig:real_01_object_trajectory}. The object position is tracked accurately, with a maximum error of \SI{6}{\milli\meter} and a mean error of \SI{3.7}{\milli\meter}. Orientation tracking exhibits a maximum error of \SI{1.01}{\degree} and a mean error of \SI{0.5}{\degree}. Overall, the tracking performance is satisfactory in view of the heavy load and the fast motions. Further improvements could be achieved through enhanced object parameter and state estimation and refined controller tuning.
Notably, the \SI{16}{\kilogram} payload exceeds the \SI{7}{\kilogram} nominal capacity of each individual arm by more than a factor of two, yet the system achieves accurate tracking with position errors below \SI{6}{\milli\meter} while satisfying all kinematic and dynamic constraints.

Figure \ref{fig:real_01_joint_trajectory} shows the joint trajectories and torques for selected robot axes.
The joint references satisfy the \gls{topp} limits on velocity, acceleration, jerk, and torque.
The measured trajectories track the references accurately, with only minor deviations.
The internal Sunrise controller overlays the desired torques $\jt^\mathrm{des}$ and applies adapted torques $\jt$.
Since the desired and applied torques match closely, the internal controller has minimal impact on system performance. Only in the first part of the trajectory, commanded torques deviate slightly from the references. This also results in an error in the internal wrench, which is shown in \figref{fig:real_01_internal_wrench}.
The desired internal wrenches increase when torque limits are reached, with the desired internal force $f_y$ reaching its limit.

Figure \ref{fig:real_01_cpu_time} shows the controller computation time. Real time feasibility is given, as the controller operates at \SI{1}{\kilo\hertz} and requires approximately \SI{0.3}{\milli\second} to compute the control inputs, with the \gls{hqp} consuming the majority of this time.
The first \gls{qp} \eqref{eq:optimization_system_acc} is larger and therefore requires more computation time than the second \gls{qp} \eqref{eq:optimization_TorqueWrench}.

%% file: 01_content/conclusion.tex
\section{Conclusion}\label{sec:conclusion}

This paper presented an integrated framework for time-optimal cooperative manipulation with dual-arm robots.
The approach extends the TOPP3-RA method to redundantly actuated systems by incorporating internal wrenches as optimization variables, enabling higher path accelerations under torque constraints.
Actuation redundancy is resolved through an HQP-based controller that enforces joint torque limits and bounds internal wrenches during trajectory tracking.

Experimental validation on a dual-arm system with two KUKA LBR iiwa 7 R800 manipulators demonstrated the manipulation of a \SI{16}{\kilogram} payload, exceeding the nominal capacity of both arms, while satisfying velocity, acceleration, jerk and torque constraints, a scenario that, to the best of the authors' knowledge, has not been experimentally demonstrated with integrated jerk-constrained time-optimal planning and torque-constrained control in prior work.
The results confirm the effectiveness of the proposed framework in exploiting the enhanced payload capabilities of cooperative manipulation.

Minor discrepancies between reference and experimental results were observed, primarily attributed to model inaccuracies and sensor noise.
Future work will focus on improving model fidelity, implementing robust estimation for uncertain object properties, and extending the framework to more complex manipulation scenarios.

%% file: main.bbl
\begin{thebibliography}{10}
\providecommand{\url}[1]{#1}
\csname url@samestyle\endcsname
\providecommand{\newblock}{\relax}
\providecommand{\bibinfo}[2]{#2}
\providecommand{\BIBentrySTDinterwordspacing}{\spaceskip=0pt\relax}
\providecommand{\BIBentryALTinterwordstretchfactor}{4}
\providecommand{\BIBentryALTinterwordspacing}{\spaceskip=\fontdimen2\font plus
\BIBentryALTinterwordstretchfactor\fontdimen3\font minus
  \fontdimen4\font\relax}
\providecommand{\BIBforeignlanguage}[2]{{%
\expandafter\ifx\csname l@#1\endcsname\relax
\typeout{** WARNING: IEEEtran.bst: No hyphenation pattern has been}%
\typeout{** loaded for the language `#1'. Using the pattern for}%
\typeout{** the default language instead.}%
\else
\language=\csname l@#1\endcsname
\fi
#2}}
\providecommand{\BIBdecl}{\relax}
\BIBdecl

\bibitem{Sherwani2020}
F.~Sherwani, M.~M. Asad, and B.~Ibrahim, ``{C}ollaborative {R}obots and
  {I}ndustrial {R}evolution 4.0 ({IR} 4.0),'' in \emph{Proc.\ of International
  Conference on Emerging Trends in Smart Technologies (ICETST)}, March 2020,
  pp. 168--172.

\bibitem{Matheson2019}
E.~Matheson, R.~Minto \emph{et~al.}, ``{H}uman--{R}obot {C}ollaboration in
  {M}anufacturing {A}pplications: {A} {R}eview,'' \emph{Robotics}, vol.~8,
  no.~4, 2019.

\bibitem{SinaMirrazaviSalehian2016}
S.~S.~M. Salehian, N.~Figueroa, and A.~Billard, ``Coordinated multi-arm motion
  planning: Reaching for moving objects in the face of uncertainty,'' in
  \emph{Proc.\ of Robotics: Science and Systems}, AnnArbor, Michigan, Jun.
  2016.

\bibitem{Dio2023}
M.~Dio, A.~Völz, and K.~Graichen, ``{C}ooperative {D}ual-{A}rm {C}ontrol for
  {H}eavy {O}bject {M}anipulation based on {H}ierarchical {Q}uadratic
  {P}rogramming,'' in \emph{Proc.\ of International Conference on Intelligent
  Robots and Systems}, Detroit, USA, 2023, pp. 643--648.

\bibitem{Reiter2018}
A.~Reiter, A.~Müller, and H.~Gattringer, ``{O}n {H}igher {O}rder {I}nverse
  {K}inematics {M}ethods in {T}ime-{O}ptimal {T}rajectory {P}lanning for
  {K}inematically {R}edundant {M}anipulators,'' \emph{IEEE Transactions on
  Industrial Informatics}, vol.~14, no.~4, pp. 1681--1690, 2018.

\bibitem{Dio2024}
M.~Dio, K.~Graichen, and A.~Völz, ``Time-optimal path parameterization for
  cooperative multi-arm robotic systems with third-order constraints,'' in
  \emph{Proc.\ of International Conference on Intelligent Robots and Systems},
  Abu Dhabi, UAE, 2024, pp. 3043--3048.

\bibitem{Dio2025}
M.~Dio, A.~Wahrburg \emph{et~al.}, ``{Time}-{Optimal} {Path} {Parameterization}
  with {Viscous} {Friction} and {Jerk} {Constraints} based on {Reachability}
  {Analysis},'' in \emph{Proc.\ of International Conference on Intelligent
  Robots and Systems}, Hangzhou, China, 2025, pp. 5848--5854.

\bibitem{Bobrow1985}
J.~Bobrow, S.~Dubowsky, and J.~Gibson, ``{T}ime-{O}ptimal {C}ontrol of
  {R}obotic {M}anipulators {A}long {S}pecified {P}aths,'' \emph{The
  International Journal of Robotics Research}, vol.~4, no.~3, pp. 3--17, 1985.

\bibitem{Pham2014}
Q.-C. Pham, ``{A} {G}eneral, {F}ast, and {R}obust {I}mplementation of the
  {T}ime-{O}ptimal {P}ath {P}arameterization {A}lgorithm,'' \emph{Transactions
  on Robotics}, vol.~30, no.~6, pp. 1533--1540, 2014.

\bibitem{Liu1991}
L.~Yun-Hui and S.~Arimoto, ``{M}inimum-{T}ime {T}rajectory {P}lanning for
  {M}ultiple {M}anipulators {H}andling an {O}bject with {G}eometric {P}ath
  {C}onstraints,'' in \emph{Proc.\ of International Conference on Intelligent
  Robots and Systems}, Osaka, Japan, 1991.

\bibitem{PhamStasse2015}
Q.-C. Pham and O.~Stasse, ``{T}ime-{O}ptimal {P}ath {P}arameterization for
  {R}edundantly {A}ctuated {R}obots: {A} {N}umerical {I}ntegration
  {A}pproach,'' \emph{Transactions on Mechatronics}, vol.~20, no.~6, pp.
  3257--3263, 2015.

\bibitem{PhamTOPP32017}
H.~Pham and Q.-C. Pham, ``{O}n the {S}tructure of the {T}ime-{O}ptimal {P}ath
  {P}arameterization {P}roblem with {T}hird-{O}rder {C}onstraints,'' in
  \emph{Proc.\ of International Conference on Robotics and Automation},
  Singapore, 2017.

\bibitem{Verscheure2008}
D.~Verscheure, B.~Demeulenaere \emph{et~al.}, ``{P}ractical {T}ime-{O}ptimal
  {T}rajectory {P}lanning for {R}obots: {A} {C}onvex {O}ptimization
  {A}pproach,'' \emph{IEEE Transactions on Automatic Control}, vol.~14, 2008.

\bibitem{Debrouwere2013}
F.~Debrouwere, W.~Van~Loock \emph{et~al.}, ``{T}ime-{O}ptimal {P}ath
  {F}ollowing for {R}obots with {T}rajectory {J}erk {C}onstraints using
  {S}equential {C}onvex {P}rogramming,'' in \emph{Proc.\ of International
  Conference on Robotics and Automation}, Karlsruhe, Germany, 2017, pp.
  1916--1921.

\bibitem{Ji2024}
C.~Ji, Z.~Zhang \emph{et~al.}, ``A convex optimization method to time-optimal
  trajectory planning with jerk constraint for industrial robotic
  manipulators,'' \emph{Transactions on Automation Science and Engineering},
  vol.~21, no.~4, pp. 7629--7646, 2024.

\bibitem{Huang2023}
L.~Huang and X.~Jing, ``Fast asymptotically optimal path planning in dynamic,
  uncertain environments,'' in \emph{Proc.\ of International Conference on
  Intelligent Robots and Systems}, Detroit, USA, 2023, pp. 2846--2852.

\bibitem{Lee2024}
J.-E. Lee, A.~Bylard \emph{et~al.}, ``On the performance of jerk-constrained
  time-optimal trajectory planning for industrial manipulators,'' in
  \emph{Proc.\ of International Conference on Robotics and Automation},
  Yokohama, Japan, 2024, pp. 9772--9778.

\bibitem{Kaserer2019}
D.~Kaserer, H.~Gattringer, and A.~Müller, ``{N}early {O}ptimal {P}ath
  {F}ollowing {W}ith {J}erk and {T}orque {R}ate {L}imits {U}sing {D}ynamic
  {P}rogramming,'' \emph{Transactions on Robotics}, vol.~35, no.~2, pp.
  521--528, 2019.

\bibitem{Kaserer2020}
D.~Kaserer, H.~Gattringer, and A.~Mueller, ``{T}ime {O}ptimal {M}otion
  {P}lanning and {A}dmittance {C}ontrol for {C}ooperative {G}rasping,''
  \emph{IEEE Robotics and Automation Letters}, vol.~5, no.~2, pp. 2216--2223,
  Apr. 2020.

\bibitem{PhamRA2017}
H.~Pham and Q.-C. Pham, ``A {N}ew {A}pproach to {T}ime-{O}ptimal {P}ath
  {P}arameterization based on {R}eachability {A}nalysis,'' \emph{Transactions
  on Robotics}, vol.~34, no.~3, pp. 645--659, 2018.

\bibitem{Smith2012}
C.~Smith, Y.~Karayiannidis \emph{et~al.}, ``Dual arm manipulation---a survey,''
  \emph{Robotics and Autonomous Systems}, vol.~60, no.~10, pp. 1340--1353,
  2012.

\bibitem{Abbas2023}
M.~Abbas, J.~Narayan, and S.~K. Dwivedy, ``A systematic review on cooperative
  dual-arm manipulators: modeling, planning, control, and vision strategies,''
  \emph{International Journal of Intelligent Robotics and Applications},
  vol.~7, no.~4, pp. 683--707, Jul. 2023.

\bibitem{Khatib1987}
O.~Khatib, ``A unified approach for motion and force control of robot
  manipulators: The operational space formulation,'' \emph{IEEE Journal on
  Robotics and Automation}, vol.~3, no.~1, pp. 43--53, 1987.

\bibitem{Williams1993}
D.~Williams and O.~Khatib, ``The virtual linkage: A model for internal forces
  in multi-grasp manipulation,'' in \emph{Proc.\ of International Conference on
  Robotics and Automation}, Atlanta, USA, 1993, pp. 1025--1030.

\bibitem{Walker1991}
I.~D. Walker, R.~A. Freeman, and S.~I. Marcus, ``Analysis of motion and
  internal loading of objects grasped by multiple cooperating manipulators,''
  \emph{The International Journal of Robotics Research}, vol.~10, no.~4, pp.
  396--409, Aug. 1991.

\bibitem{Chung2005}
J.~H. Chung, B.-J. Yi, and W.~K. Kim, ``Analysis of internal loading at
  multiple robotic systems,'' \emph{Journal of Mechanical Science and
  Technology}, vol.~19, no.~8, pp. 1554--1567, Aug. 2005.

\bibitem{Erhart2015}
S.~Erhart and S.~Hirche, ``Internal force analysis and load distribution for
  cooperative multi-robot manipulation,'' \emph{Transactions on Robotics},
  vol.~31, no.~5, pp. 1238--1243, 2015.

\bibitem{KIKUCHI19842349}
H.~Kikuchi, T.~Niinomi \emph{et~al.}, ``Heavy parts assembly by coordinative
  control of robot and balancing manipulator,'' \emph{IFAC Proceedings
  Volumes}, vol.~17, no.~2, pp. 2349--2354, 1984.

\bibitem{Uchiyama1988}
M.~Uchiyama and Y.~Nakamura, ``Symmetric hybrid position/force control of two
  cooperating robot manipulators,'' in \emph{Proc.\ of International Workshop
  on Intelligent Robots}, 1988, pp. 515--520.

\bibitem{Hogan1985}
N.~Hogan, ``Impedance control: An approach to manipulation: {P}art
  {I}---{T}heory,'' \emph{Journal of Dynamic Systems, Measurement, and
  Control}, vol. 107, no.~1, pp. 1--7, 1985.

\bibitem{Caccavale2008}
F.~Caccavale, P.~Chiacchio \emph{et~al.}, ``{Six-DOF impedance control of
  dual-arm cooperative manipulators},'' \emph{IEEE/ASME Transactions on
  Mechatronics}, vol.~13, no.~5, pp. 576--586, 2008.

\bibitem{Erhart2016}
S.~Erhart and S.~Hirche, ``{Model and Analysis of the Interaction Dynamics in
  Cooperative Manipulation Tasks},'' \emph{Transactions on Robotics}, vol.~32,
  no.~3, pp. 672--683, 2016.

\bibitem{Verginis2023}
C.~K. Verginis, D.~Zelazo, and D.~V. Dimarogonas, ``Cooperative manipulation
  via internal force regulation: A rigidity theory perspective,'' \emph{IEEE
  Transactions on Control of Network Systems}, vol.~10, no.~3, pp. 1222--1233,
  2023.

\bibitem{Nakamura1987}
Y.~Nakamura, H.~Hanafusa, and T.~Yoshikawa, ``Task-priority based redundancy
  control of robot manipulators,'' \emph{The International Journal of Robotics
  Research}, vol.~6, no.~2, pp. 3--15, 1987.

\bibitem{Siciliano1991}
B.~Siciliano and J.-J.~E. Slotine, ``A general framework for managing multiple
  tasks in highly redundant robotic systems,'' in \emph{Proc.\ of International
  Conference on Advanced Robotics}, 1991, pp. 1211--1216.

\bibitem{Sentis2005}
L.~Sentis and O.~Khatib, ``Synthesis of whole-body behaviors through
  hierarchical control of behavioral primitives,'' \emph{International Journal
  of Humanoid Robotics}, vol.~2, no.~4, pp. 505--518, 2005.

\bibitem{Kanoun2011}
O.~Kanoun, F.~Lamiraux, and P.-B. Wieber, ``Kinematic control of redundant
  manipulators: Generalizing the task-priority framework to inequality tasks,''
  \emph{Transactions on Robotics}, vol.~27, no.~4, pp. 785--792, 2011.

\bibitem{Escande2014}
A.~Escande, N.~Mansard, and P.-B. Wieber, ``Hierarchical quadratic programming:
  Fast online humanoid-robot motion generation,'' \emph{The International
  Journal of Robotics Research}, vol.~33, no.~7, pp. 1006--1028, may 2014.

\bibitem{carpentier:hal-03271811}
J.~Carpentier, R.~Budhiraja, and N.~Mansard, ``{Proximal and Sparse Resolution
  of Constrained Dynamic Equations},'' in \emph{Proc.\ of Robotics: Science and
  Systems}, Virtual, Jul. 2021.

\bibitem{wanner1996solving}
G.~Wanner and E.~Hairer, \emph{Solving ordinary differential equations
  II}.\hskip 1em plus 0.5em minus 0.4em\relax Springer Berlin Heidelberg New
  York, 1996, vol. 375.

\bibitem{Hoffman2018}
E.~M. Hoffman, A.~Laurenzi \emph{et~al.}, ``Multi-priority {C}artesian
  impedance control based on quadratic programming optimization,'' in
  \emph{Proc.\ of International Conference on Robotics and Automation},
  Brisbane, Australia, 2018, pp. 309--315.

\bibitem{diamond2016cvxpy}
S.~Diamond and S.~Boyd, ``{CVXPY}: {A} {P}ython-embedded modeling language for
  convex optimization,'' \emph{Journal of Machine Learning Research}, vol.~17,
  no.~83, pp. 1--5, 2016.

\end{thebibliography}
